\documentclass[letterpaper, 10 pt, conference]{ieeeconf}  

\usepackage{graphicx}
\graphicspath{{img}}
\usepackage{placeins}

\usepackage{fancyhdr}
\fancypagestyle{withfooter}{
  
  \fancyfoot[C]{\footnotesize Accepted to the IEEE ICRA Workshop on Field Robotics 2024}
}

\IEEEoverridecommandlockouts                              

\title{\LARGE \bf
Challenges in automatic and selective plant-clearing
}

\author{Fabrice Mayran de Chamisso$^{1}$, Loïc Cotten$^{2}$,\\
Valentine Dhers, Thomas Lompech, Florian Seywert and Arnaud Susset$^{3}$
\thanks{$^{1}$Fabrice Mayran de Chamisso is with Université Paris Saclay, CEA, LIST, 91120 Palaiseau, France
        {\tt\small fabrice.mayran-de-chamisso@cea.fr}}%
\thanks{$^{2}$Loïc Cotten is with Alliance Forêts Bois
	{\tt\small loic.cotten@alliancefb.fr}}%
\thanks{$^{3}$V. Dhers, T. Lompech, F. Seywert and A. Susset are with R\&D Vision, 64 rue Bourdignon, 94100 St Maur des Fossés, France
	{\tt\small arnaud.susset@rd-vision.com}}%
}

\begin{document}

\maketitle
\thispagestyle{withfooter}
\pagestyle{withfooter}

\begin{abstract}

With the advent of multispectral imagery and AI, there have been numerous works on automatic plant segmentation for purposes such as counting, picking, health monitoring, localized pesticide delivery, etc. In this paper, we tackle the related problem of automatic and selective plant-clearing in a sustainable forestry context, where an autonomous machine has to detect and avoid specific plants while clearing any weeds which may compete with the species being cultivated. Such an autonomous system requires a high level of robustness to weather conditions, plant variability, terrain and weeds while remaining cheap and easy to maintain. We notably discuss the lack of robustness of spectral imagery, investigate the impact of the reference database's size and discuss issues specific to AI systems operating in uncontrolled environments.

\end{abstract}

\section{Introduction - challenges}

\begin{figure}
\begin{center}
\includegraphics[width=0.9\columnwidth]{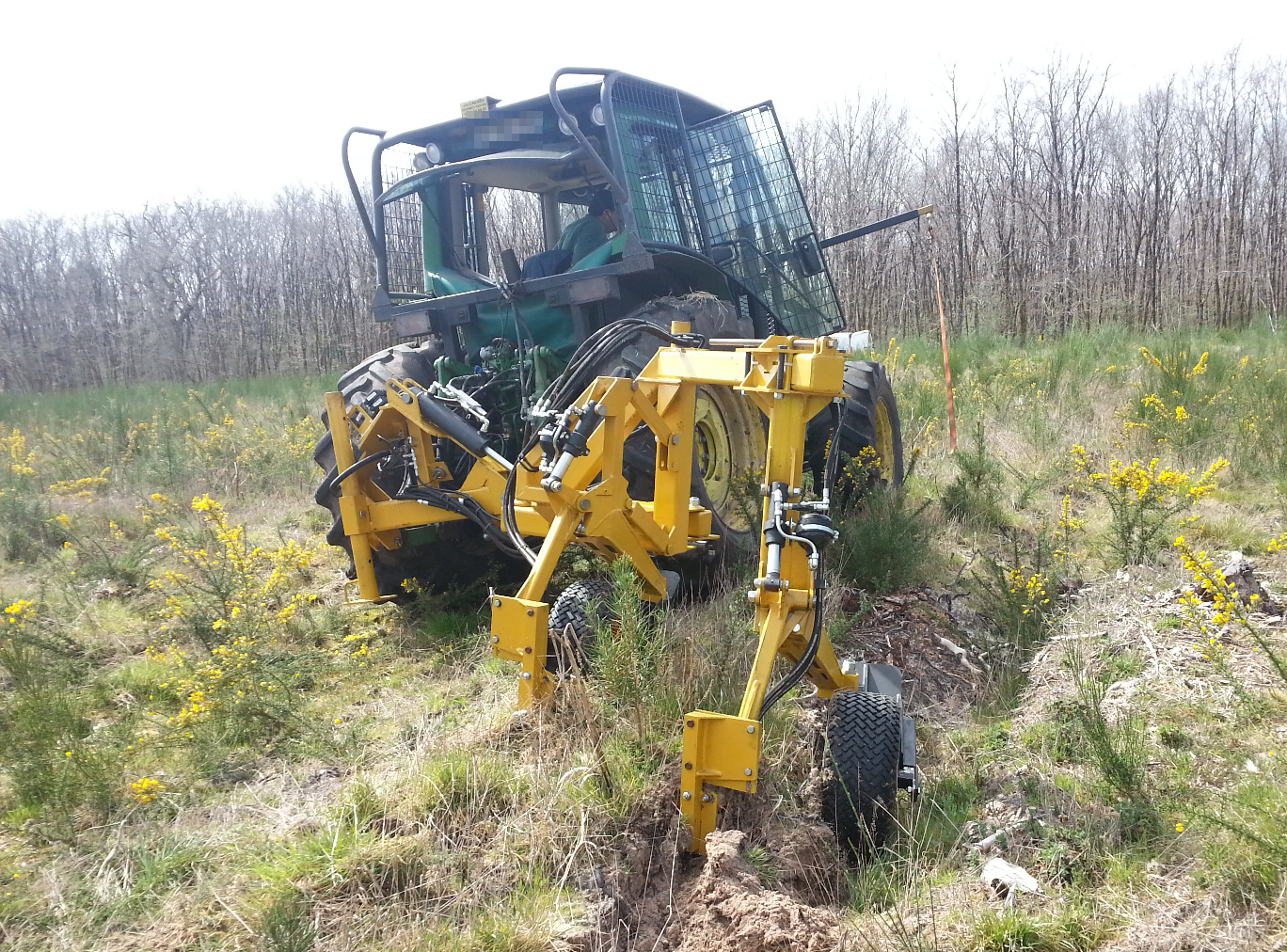}
\includegraphics[width=0.9\columnwidth]{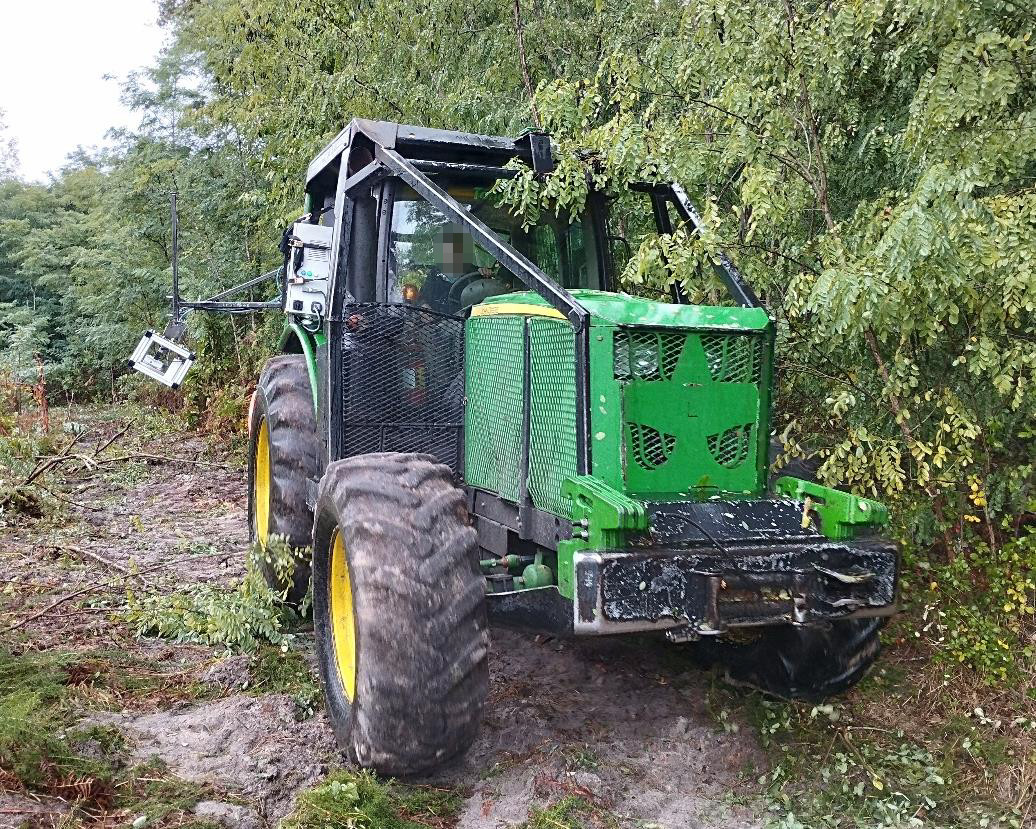}
\caption{Top: Typical existing tractor with retractable cutting tool. Bottom: Ruggedized acquisition system attached to a tractor for data collection.}\label{figtractor}
\end{center}
\end{figure}

Reforestation and sustainable forestry is a massive worldwide challenge, which is important for ecosystems in addition to being an income source. However, manual plant clearing is a time-consuming task, which is usually carried by a team of workers equipped with some form of brushcutter or sitting on a tractor with a retractable arm such as the one shown on Figure \ref{figtractor}. The main issue with brushcutters is the arduousness of the job, possibly causing MSDs. Also, clearing some plants while avoiding others requires constant attention from the operator. The retractable arm tractor requires even more attention from the driver, since fields where saplings are planted are not level, notably exhibiting rocks, tree stumps and small ponds. Consequently, the driver has to focus on driving, and can't pilot the retractable arm reliably.

\begin{figure}
\begin{center}
\includegraphics[width=1.0\columnwidth]{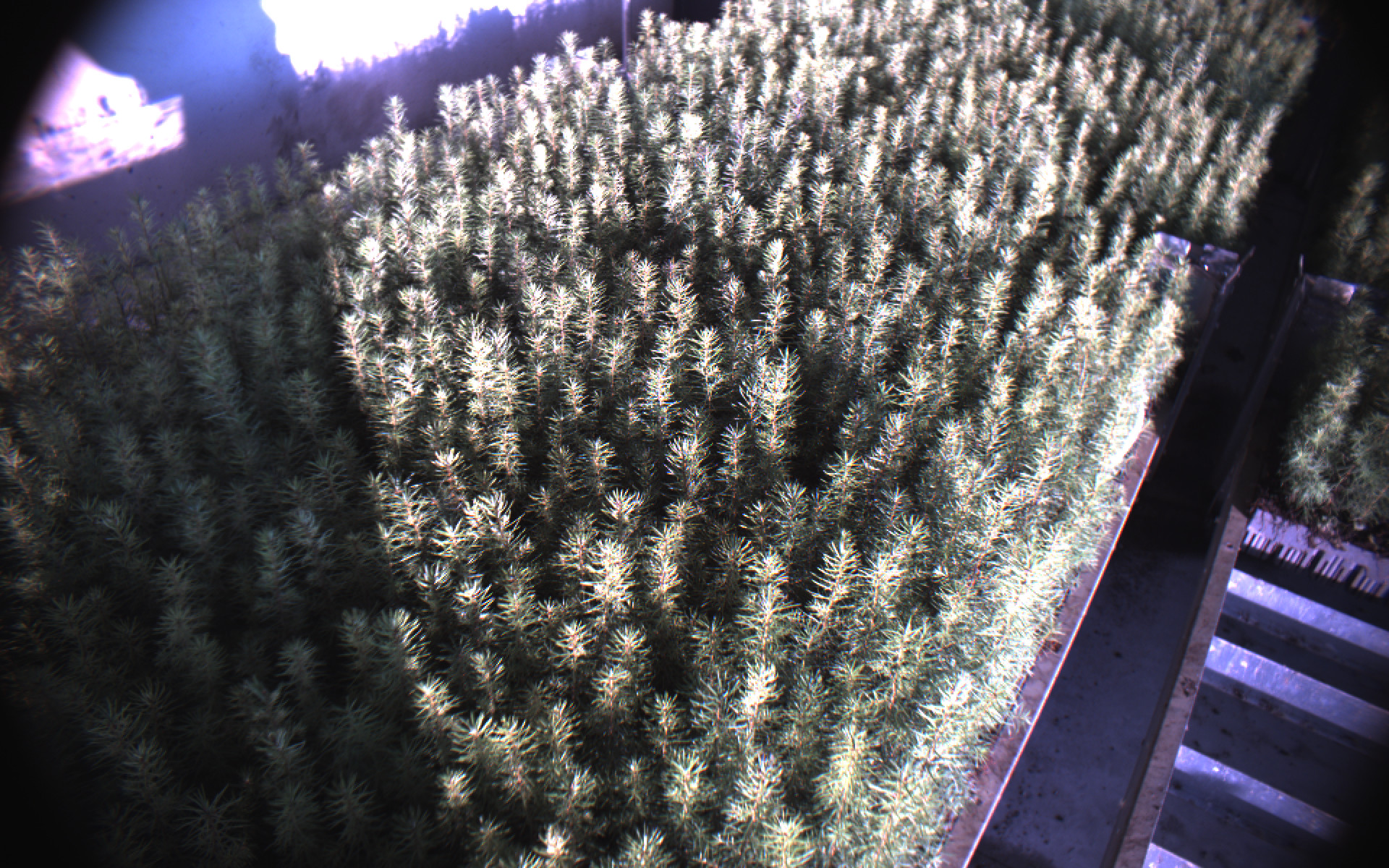}
\caption{Saplings before planting measure between a few centimeters and a few dozens of centimeters.}\label{figsaplings}
\end{center}
\end{figure}

In theory, saplings (see Figure \ref{figsaplings}) are evenly spaced and planted in lines. However, due to terrain obstacles such as rocks or stumps, this regular pattern should not be taken for granted. As a result, the position of saplings is uncertain, even though they roughly form a line.

Our goal is to first develop a semi-autonomous system piloting the retractable arm of a tractor such as that of Figure \ref{figtractor}. The operator would focus on following the approximative line of saplings while the autonomous cutting tool would retract and extend as needed. As a second step, a fully autonomous machine could be developed, integrating autonomous driving and obstacle avoidance. We are only considering a set of coniferous species (\textit{Pinus taeda}, \textit{Pinus pinaster}, \textit{Pseudotsuga menziesii} and \textit{Larix}), although the challenges and solutions discussed in this paper should apply to any tree or plant species.

The rest of this paper goes as follows: first, we discuss the choice of sensors, notably including multispectral imagery as a proxy to segment saplings. Then, we explore sapling segmentation using AI detection and segmentation with a custom fuzzy segmentation approach and an off-the-shelf YOLO(v8) network \cite{ultralytics2021}. Finally, we discuss integration challenges such as temporal consistency and all-season all-weather operation.

\section{Vision system design}

\begin{figure}
\begin{center}
\includegraphics[width=\columnwidth]{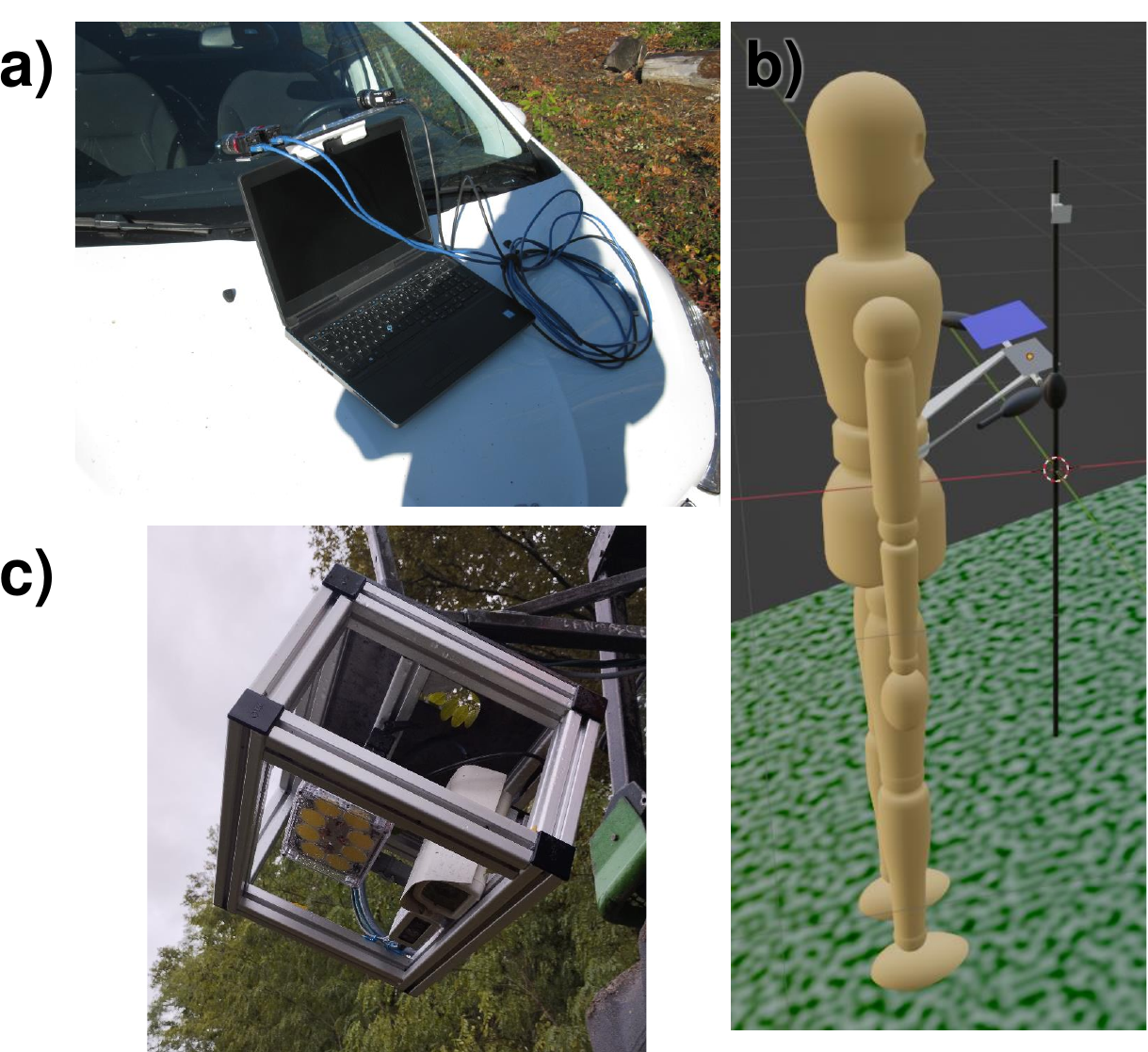}
\caption{Evolution of acquisition systems. a) initial acquisition system, consisting of two manually triggered IMEC multispectral cameras and one manually triggered industrial global shutter FLIR camera with a SONY IMX sensor. Multispectral cameras have a resolution of 512x256x16 in the visible spectrum and 409x216x25 in the near infrared spectrum. This setup was complemented with a customer grade compact camera and a smartphone sensor. b) database construction system idea, with a sensor mounted on a belt-worn device, a feedback screen and a control panel. In the end, this system was simplified to a wrist-strapped $10$inch monitor, a hand-held camera and a single button to take a picture or a video. c) the ruggedized acquisition system used for all-terrain all-season operation, with a builtin illumination system for low-light operation.}\label{figacqui}
\end{center}
\end{figure}

In order to maximize transferability of plant detection algorithms, it is best to use the same sensor for the acquisition of a database used to train and validate algorithms as well as for the final application. Figure \ref{figacqui} presents different stages of acquisition system from less practical and rugged to almost finalized. The rugged system was mounted on a tractor as shown on Figure \ref{figtractor}.

\subsection{Sensor choice: multispectral versus classical imagery}

\textbf{Spectral imagery}
In recent years, multispectral imagery with a few selected channels has been considered one of the, if not the best proxy for vegetation segmentation. Indeed, many components of plants such as water and pigments (cholorophylle, carotene, xantophylls, anthocyanins, etc.) \cite{Kim2018} have a localized spectral signature. The water content of a plant, for instance, can be determined in the medium wave infrared ($900-2500nm$) spectrum \cite{Adam2010}. The meta-study by Fassnacht et al. \cite{Fassnacht2016} reveals that literature mostly studies the $450-550nm$ (visible) and $650-700nm$ (red edge) frequency bands. These ranges are used by \cite{Dalponte2012} to perform tree species classification from aerial imagery. \cite{Santos2010} uses $400-800nm$, $520-550nm$ and $680-700nm$ for health monitoring of pines.

Interestingly, hyperspectral imagery using many spectral channels only bring marginal improvements to classification/segmentation performances \cite{Fassnacht2016} compared to multispectral imagery using just a few channels.

\begin{figure}
\begin{center}
\includegraphics[width=0.5\columnwidth]{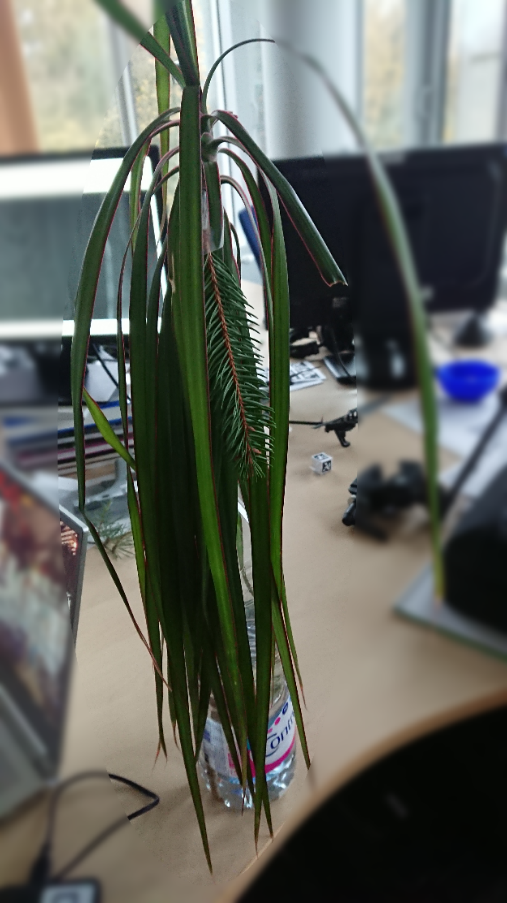}
\includegraphics[width=1.0\columnwidth]{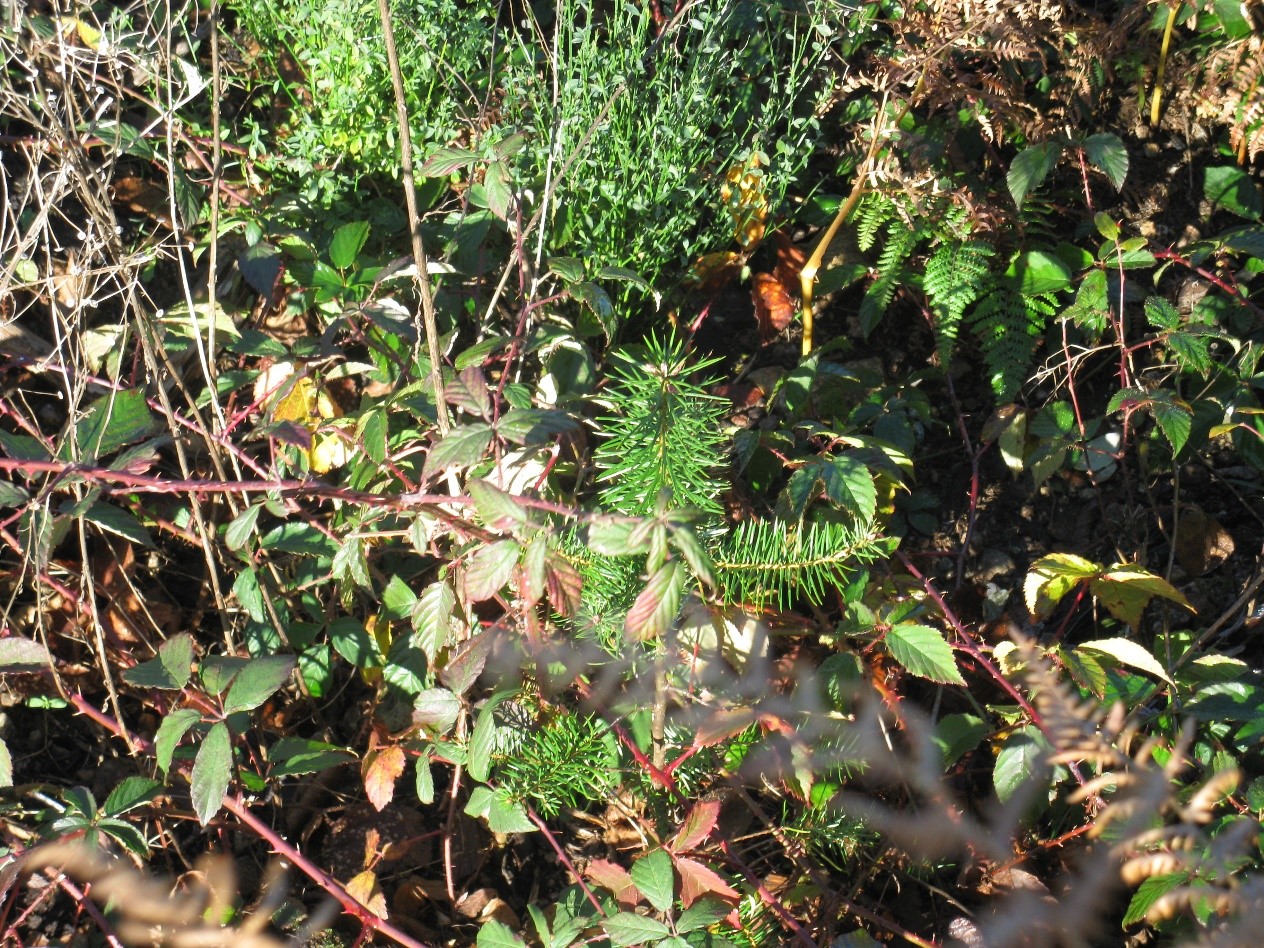}
\caption{In the laboratory vs reality. Top: in a controlled environment, spectral segmentation shows promising results. Bottom: in reality, intraclass variability is similar to interclass variability, with uncontrolled lighting.}\label{figfaketree}
\end{center}
\end{figure}

All in all, spectral imagery seemed promising for coniferous detection among other plants. We used simple classification methods such as Spectral Angular Mapping, which consists in attaching each region of an image to a class based on the cosine similarity or dot product between spectra. Although this worked in the laboratory (Figure \ref{figfaketree}), we could never achieve satisfactory accuracy on real data, since intraclass variability was similar to interclass variability. Furthermore, in reality, lighting is not controlled and possibly fast-changing while spectral analysis tools are very dependent on normalization.

Spectral imagery has other issues: the cost of the sensors is higher, especially if a very narrow non-standard spectral band is required. If multiple bands are required, the sensor's resolution has to be lowered to allow for simultaneous acquisition of all bands. Finally, in addition to environmental factors such as dust or humidity which may change the color of a plant, some plots have to be treated with deer repellent which gives saplings a distinct white tint and completely changes their spectral signature.

\begin{figure*}
\begin{center}
\includegraphics[width=1.0\textwidth]{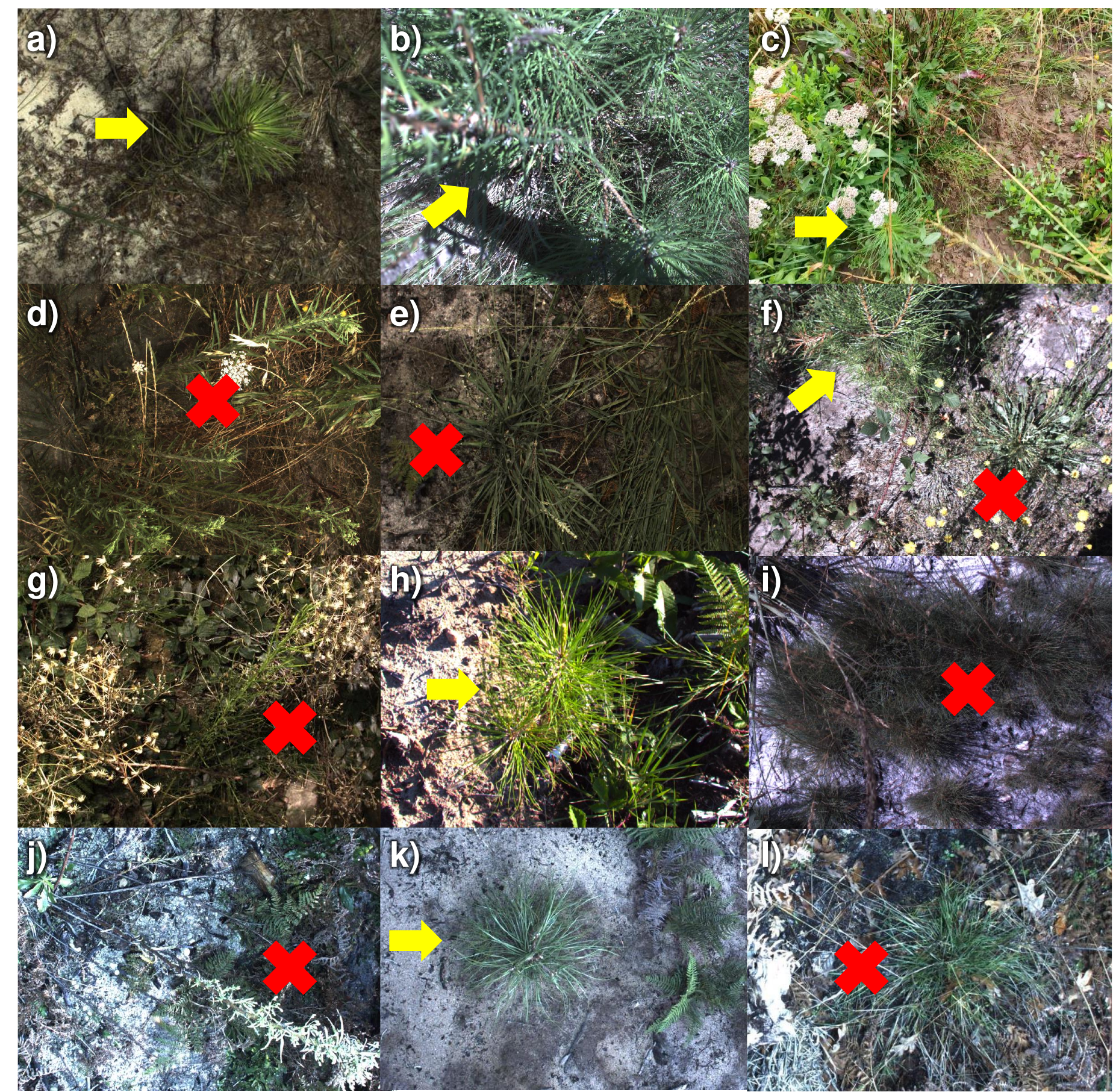}
\caption{A few examples from the collected database. Actual coniferous samples are shown with a yellow arrow while plants closely resembling coniferous species are shown with a red cross.}\label{figexdataset}
\end{center}
\end{figure*}

While collecting samples for spectral segmentation, we confirmed the intuition that coniferous species are morphologically distinct from most other species due to their distinctive pointy leaves and apical bud aspect. These are more prominent on a high resolution RGB sensor then on a lower resolution multispectral sensor. This finding, along with the other issues of multispectral imagery, incentivized us to switch to classical RGB imagery using a high resolution global shutter industrial camera. We were especially careful to include samples of \textit{ulex} ("furze"), various \textit{fabaceae} ("broom") and grass (see Figure \ref{figexdataset}) in our dataset since these species may be visually similar to coniferous species from some points of views.

\textbf{Sensor and parameters}
In order to build a man-portable system, we chose a global shutter FLIR industrial sensor with resolution $2048x1536$. It should be noted that this resolution is much larger than can be processed by existing neural networks. However, using a higher definition allows cropping the image to simulate different sensor positions and distance to the plant. It also allows zooming while performing the annotation task, which proved useful on challenging images with high occlusion rate.

\begin{figure}
\begin{center}
\includegraphics[width=1.0\columnwidth]{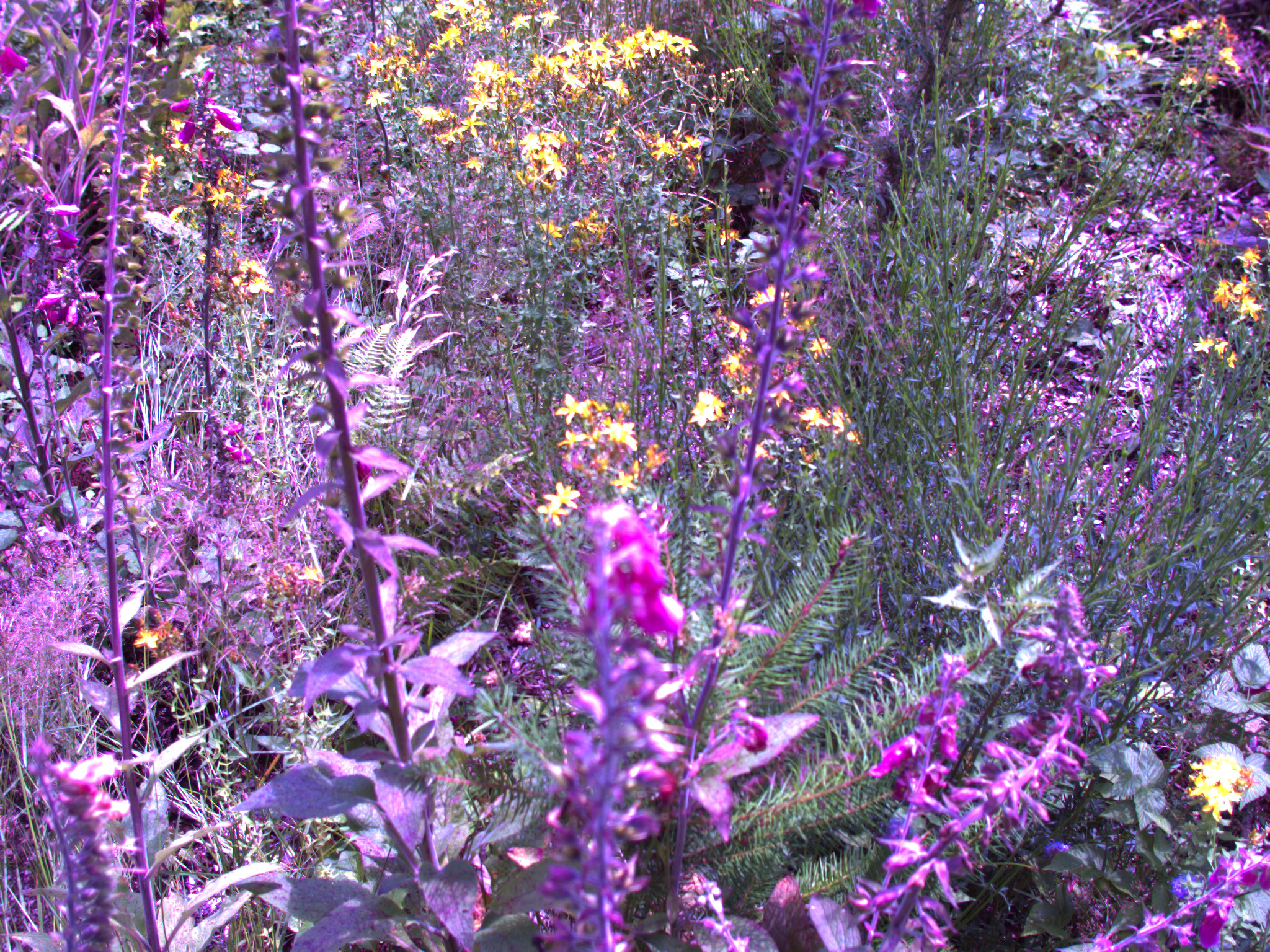}
\caption{Without fixing the white balance, sensors may produce weird looking images. There is no purple plant or soil here in reality.}\label{figpurple}
\end{center}
\end{figure}

We had issues with the white balance of the sensor, as shown on Figure \ref{figpurple}. This common problem was solved by setting a fixed white balance. Regarding focus, setting a focus once and for all is a challenge if acquisitions are performed from multiple distances or if plants are tall and one part is out of focus (Figure \ref{figexdataset} (b)). This can be solved with a diaphragm at the cost of reduced light hitting the sensor. We added external illumination to combat that as well as reduce overall lighting variability. Regarding exposure time/ISO, we found that the best looking images were obtained by letting the camera automatically set its parameters but only within a reduced adaptation range. This does not solve the issue of fast transitions between bright and dark zones, which is handled at a later stage of the pipeline, using temporal stabilization as explained in section \ref{sectempstab}.

Cameras in outside environments will get dirty. It is possible to detect this phenomenon with approaches such as \cite{Uricar2019} in addition to requiring the operator to monitor a feedback screen every now and then and to perform visual inspection of the sensor at fixed intervals or when specific events occur (collision, heavy rainfall, etc.).

\FloatBarrier

\textbf{Annotation and detection challenges}

We collected a multiple databases of sapling pictures in fields, for a total of about $15000$ images. The data collection campaign lasted for multiple months in order to capture as much variability as possible. Captured variability notably includes:
\begin{itemize}
\item{plant species (taeda, pinaster, menziesii, larix)}
\item{plant age (1-4 years)}
\item{weather and humidity (sunny, cloudy, changing, low and high humidity levels)}
\item{time of day and lighting conditions}
\item{season}
\item{surrounding vegetation density and nature}
\end{itemize}

Figure \ref{figexdataset} shows some examples of the collected dataset and reveals some of the challenges for annotation as well as for detection and segmentation. a) is the nominal ``easy" case of a sapling on a simple soil background. b) presents a sapling which is big enough to get out of focus. c) is an example of cluttered environments where annotation is hard. Also notice how the shades of green in c) and h) are different from other images such as (b). d), e), g), i) and j) are various plants resembling saplings. f), k) and l) are some of the most difficult cases where grass is almost indistinguishable from a sapling from a top view. This can be somewhat alleviated by tilting the camera in order to never encounter top views.

\begin{figure}
\begin{center}
\includegraphics[width=1.0\columnwidth]{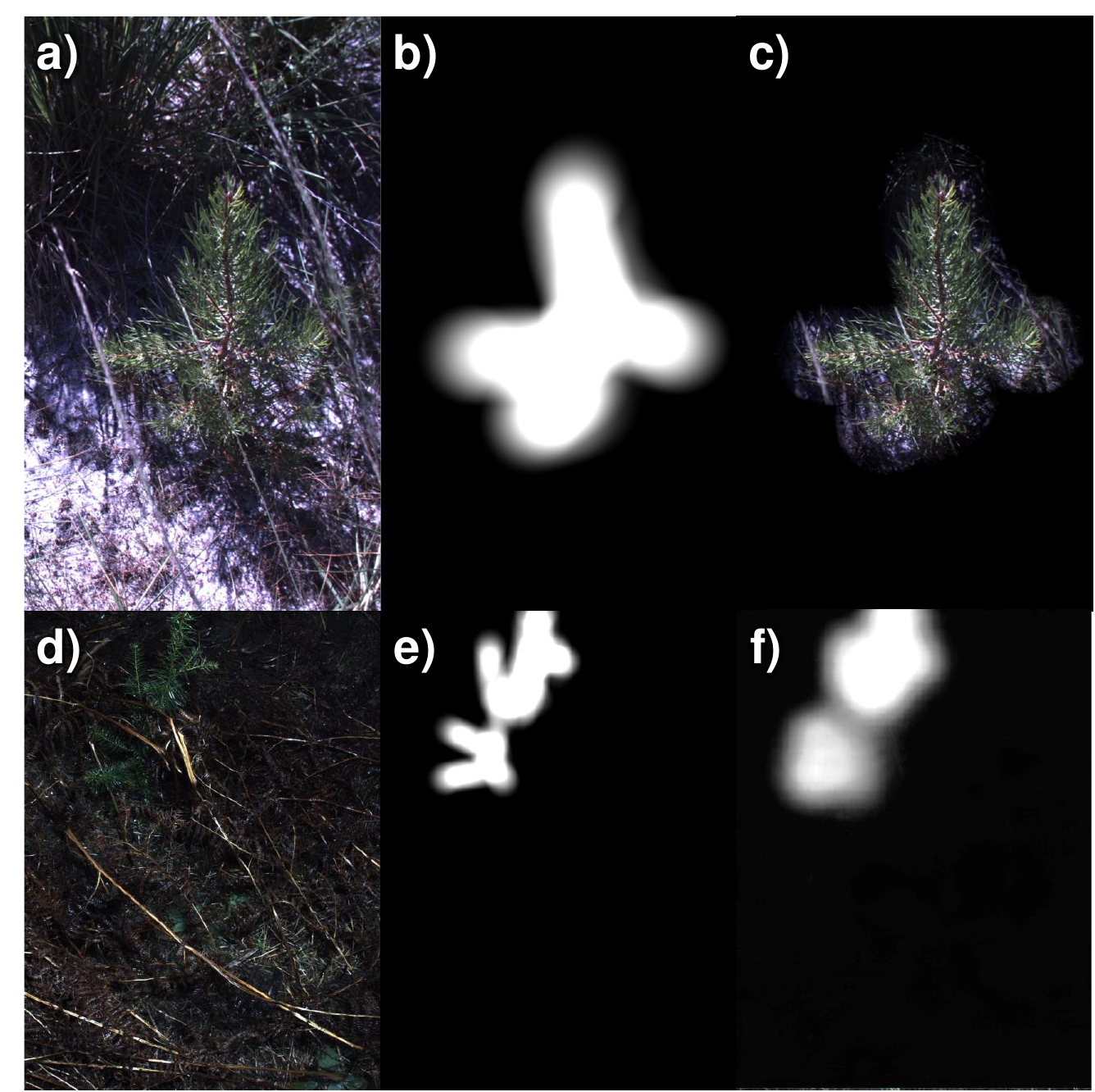}
\caption{Fuzzy annotation and learning with a U-Net++ \cite{Zhou2018}-like network. a) source sapling image. b) fuzzy ``spray can" annotation on (a). c) (b) superposed on (a). d) another sapling image. e) fuzzy annotation on (d). f) predicted fuzzy segmentation mask on (d).}\label{figfuzzy}
\end{center}
\end{figure}

We wanted to see to which extent it was possible to directly produce a segmentation mask in order to gain precision. For this purpose, we deployed a custom annotation software, producing spray-paint-like fuzzy annotations (Figure\ref{figfuzzy}). We used fuzzy annotations instead of polygons since precisely segmenting each leaf of a coniferous is intractable. Fuzzy annotations have the benefit of allowing the annotator to express their confidence in each annotation at the pixel level and at the instance level while not requiring much time. The annotation time varies between two seconds for images without a sapling to about ten seconds for images with a complex geometry sapling such as Figure \ref{figfuzzy} d).

Since some images lack contrast, our fuzzy annotation software uses an optional CLAHE filter. We found out that zooming and using the temporal consistency between successive frames due to the acquisition method of walking or driving the camera along the sapling line was required in many cases to obtain accurate annotations.

\section{Plant segmentation with deep learning}

\subsection{Data handling and learning}
The YOLO family of networks \cite{Redmon2017} is widely recognized as having a good detection performances to inference cost ratio due to their single-pass nature \cite{Zhao2019} and focus on speed. We used a YOLOV8-Large network for bounding-box-based detection, where bounding boxes were obtained from fuzzy annotations.
In parallel, we used a U-Net++\cite{Zhou2018}-like architecture tuned to $640\times 480$ RGB inputs to directly predict the fuzzy mask with an euclidean distance loss. Detections were obtained by thresholding and clustering the obtained fuzzy mask.
Both networks use heavy data augmentations to compensate for rotations, translations, scales, lighting, color balance, etc.

We tested both networks in three configurations:\\
a) using a completely separate database for train and test, with $500$ train images, $120$ test images and $120$ validation images\\
b) training on the $500$ images train database and testing on a $8500$ images test set\\
c) splitting the $8500$ image set randomly into $80\%$ train and $20\%$ test images. This last configuration has one major issue: images were acquired in sequence (with a somewhat fixed distance of about one meter between each), so the sample plant may appear in both the train and test but from a different point of view. This information leak from the train to the test set should however be minimal due to the perspective change between points of view.\\

\begin{figure}
\begin{center}
\includegraphics[width=1.0\columnwidth]{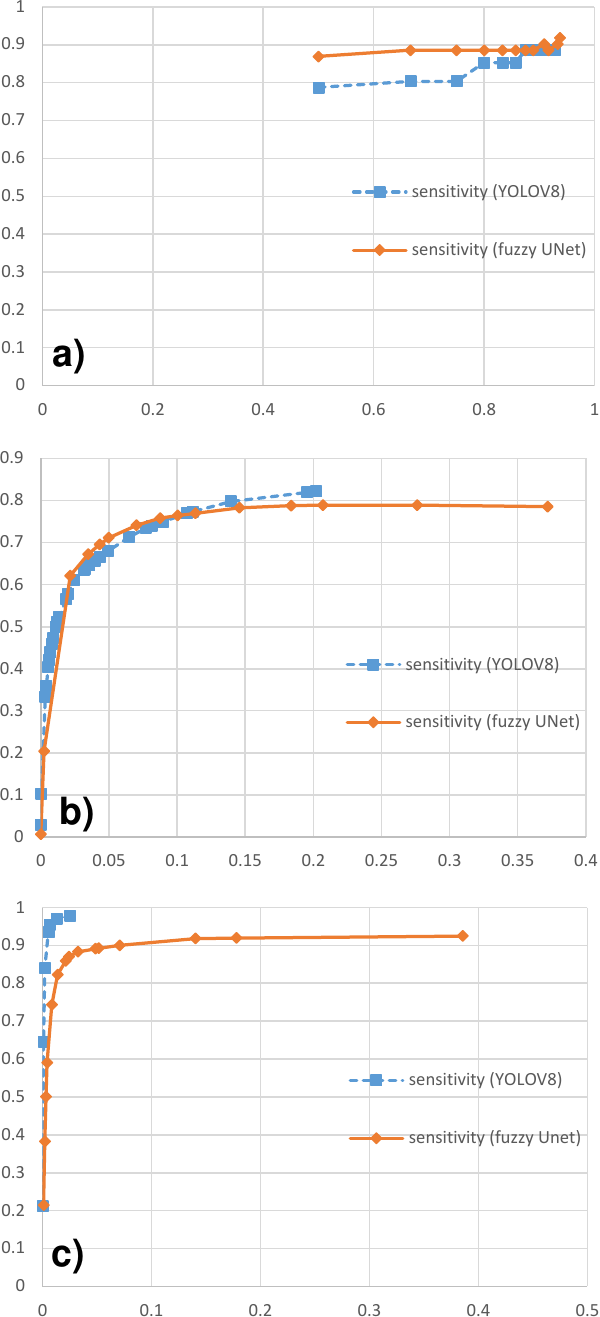}
\caption{ROC curve (true positive rate as a function of false positive rate) for all three experiments and two settings: YOLOV8 detection and fuzzy segmentation. a) limited train and test, b) limited train, large test and c) large train and test}\label{figroc}
\end{center}
\end{figure}

\begin{table}[t]
\begin{center}
\begin{tabular}{ c|c|c|c }
configuration & a & b & c\\ 
\hline
max accuracy (\%) $\uparrow$& $80.3/\mathbf{85.9}$ & $84.0/\mathbf{84.6}$ & $\mathbf{97.9}/93.0$\\
\hline
AUROC (\%) $\uparrow$& $62/\mathbf{66}$ & $\mathbf{87.2}/80.9$ & $\mathbf{98.7}/93.9$
\end{tabular}
\end{center}
\caption{Maximum accuracy and Area Under ROC curve on the three experiments a) limited train and test, b) limited train, large test and c) large train and test, for YOLOV8/Fuzzy Unet.}
\label{tabsynthetic}
\end{table}

\subsection{Results and analysis}
Figure \ref{figroc} and table \ref{tabsynthetic} show detection performances of the two neural networks tested in the three (train set, test set) configurations. Configuration a) lacks both training and test data and highly favors fuzzy segmentation. This may be attributed to the extra information provided by fuzzy annotation compared to a bounding box. It may also be due to a dependency of the YOLOV8 network on a lot of training data to train its weights, leading to overfitting when not enough data is available. We did not observe significant differences between YOLOV8 network sizes (small, medium, large, \ldots) in this experiment, meaning that network size is not the limiting factor. When increasing the amount of test data with the same small training set, the gap between both approaches narrows, with the YOLOV8 network performing better in extreme sensitivity/specificity regimes but the fuzzy annotation approach keeping an edge in terms of maximum accuracy and detection performances around a reasonable working point ($95\%$ specificity, leading to $71\%$ sensitivity). When increasing the train set size, the YOLOV8 shows its architecture benefits and performs much better than the fuzzy segmentation network, with an accuracy close to $98\%$. It is however possible that the YOLOV8 overfits much more due to the train/test split on this large dataset, which may leak information from the train to the test set. Interestingly, the YOLOV8 does not offer much tuning on its working point, not allowing trading specificity and sensitivity. The fuzzy segmentation network, despite its architectural limitations, is much more flexible on this aspect.

All in all, the fuzzy segmentation approach used is able to capture more information from the same set of ground truth data but does not scale as well on large training data, due to YOLOV8's more refined structure and layer configuration.

\section{Temporal aspects}\label{sectempstab}

\subsection{Temporal stabilization}
In order to increase detection performances, we perform temporal smoothing of the results, based on the fact that saplings are visible in multiple successive frames. A sapling is validated if it is present at a consistent position in at least $n$ captured frames. The estimated position of the sampling is computed from the image position (center of the bounding box) and tracked using the estimated motion of the tractor. The choice of $n$ is an additional proxy to tune the sensitivity/specificity working point, which partly solves the YOLOV8 network's shortcomings in terms of tuning. Performing temporal stabilization also alleviates occlusions, where the sapling is hidden beneath a larger plant. In our dataset, saplings were often much smaller than surrounding ferns, so that temporal stabilization greatly helped improve detection performances. We do not show numerical results here because tuning $n$ is very specific to the setup used (height and angle of the camera) and to the speed of the tractor.

\subsection{Tool control}
In order to retract the tractor's cutting tool appropriately, we suppose that the tractor's speed $v_t$ is always known. It can be obtained directly from the tractor or through GPS, wheel odometry, binocular SLAM, \ldots. We also suppose that the tractor moves locally in a straight line and that the cutting tool's extension and retraction times $t_r$ and $t_e$ have a fixed, known value. Each sapling detection validated by temporal stabilization is stored as an absolute position $x_s$. The retraction decision is taken at $x = x_s - t_r.v_t - \delta_x(v_t)$ and the extension decision at $x = x_s + \delta_x(v_t)$ where $\delta_x(v_t)$ is a safety margin depending on the tractor's speed.

\subsection{Lifelong learning}
Data will keep being collected during operations in order to increase the database. In particular, detections that occurred in a frame but not in the following ones may be selectively reviewed and annotated in order to increase the detection performances.

\section{Conclusion}

We demonstrated a solution for semi-automatic and selective plant-clearing, where a human driver performs the challenging task of piloting a tractor but AI is used to drive an autonomous tool. We originally intended to use multispectral imagery for sapling detection but ended up going for RGB imagery instead due to the poor discrimination performances of affordable multispectral imagery and the one-class nature of the problem where the spectrum of the background is arbitrary. We used a YOLOV8 network to perform sapling detection in addition to a custom fuzzy segmentation method based on a UNet structure. We are considering creating a YOLOV8-Unet combination to combine the advantages of the YOLOV8 (essentially, its better architecture) and the extra information captured by fuzzy annotation and segmentation. As a closing word, we would like to emphasize the fact that the sapling detection approach is only the tip of the iceberg when designing an automatic selective plant-clearing system. In the end, whether the system is viable comes down to collected data representativity, how rugged the assembly is, and whether the tractor operators are willing to use the system because it saves them effort and time.


\bibliographystyle{IEEEtran}
\bibliography{vision.bib}

\end{document}